\documentclass[11pt,a4paper]{article}
\usepackage[hyperref]{acl2020}
\usepackage{times}
\usepackage{latexsym}
\usepackage{booktabs}
\usepackage{tikz}
\usepackage{amsmath}
\usepackage{dsfont}
\usepackage{algorithmic}
\usepackage[ruled]{algorithm2e}
\usepackage{graphicx}
\usepackage{subfigure}
\usepackage{amssymb}

\usepackage{url}

\usepackage{CJK}

\aclfinalcopy 

\title{Multi-Perspective Inferrer:\\Reasoning Sentences Relationship from Holistic Perspective}

\author{Zhen Cheng, Zaixiang Zheng, Xin-Yu Dai, Shujian Huang, Jiajun Chen \\
National Key Laboratory for Novel Software Technology, Nanjing University, Nanjing, China \\
Collaborative Innovation Center of Novel Software Technology and Industrialization, Nanjing, China \\
\{chengzhen, zhengzx\}@smail.nju.edu.cn, \{daixinyu, huangsj, chenjj\}@nju.edu.cn
}

\date{}

\begin{document}
\maketitle
\begin{abstract}

Natural Language Inference (NLI) aims to determine the logic relationships (i.e., entailment, neutral and contradiction) between a pair of premise and hypothesis. Recently, the alignment mechanism effectively helps NLI by capturing the aligned parts (i.e., the similar segments) in the sentence pairs, which imply the perspective of entailment and contradiction. However, these aligned parts will sometimes mislead the judgment of neutral relations. Intuitively, NLI should rely more on multiple perspectives to form a holistic view to eliminate bias. In this paper, we propose the Multi-Perspective Inferrer (MPI), a novel NLI model that reasons relationships from multiple perspectives associated with the three relationships. The MPI determines the perspectives of different parts of the sentences via a routing-by-agreement policy and makes the final decision from a holistic view. Additionally, we introduce an auxiliary supervised signal to ensure the MPI to learn the expected perspectives. Experiments on SNLI and MultiNLI show that 1) the MPI achieves substantial improvements on the base model, which verifies the motivation of multi-perspective inference; 2) visualized evidence verifies that the MPI learns highly interpretable perspectives as expected; 3) more importantly, the MPI is architecture-free and compatible with the powerful BERT.

\end{abstract}

\section{Introduction}

\begin{figure}
    \centering
    \includegraphics[width=7.75cm]{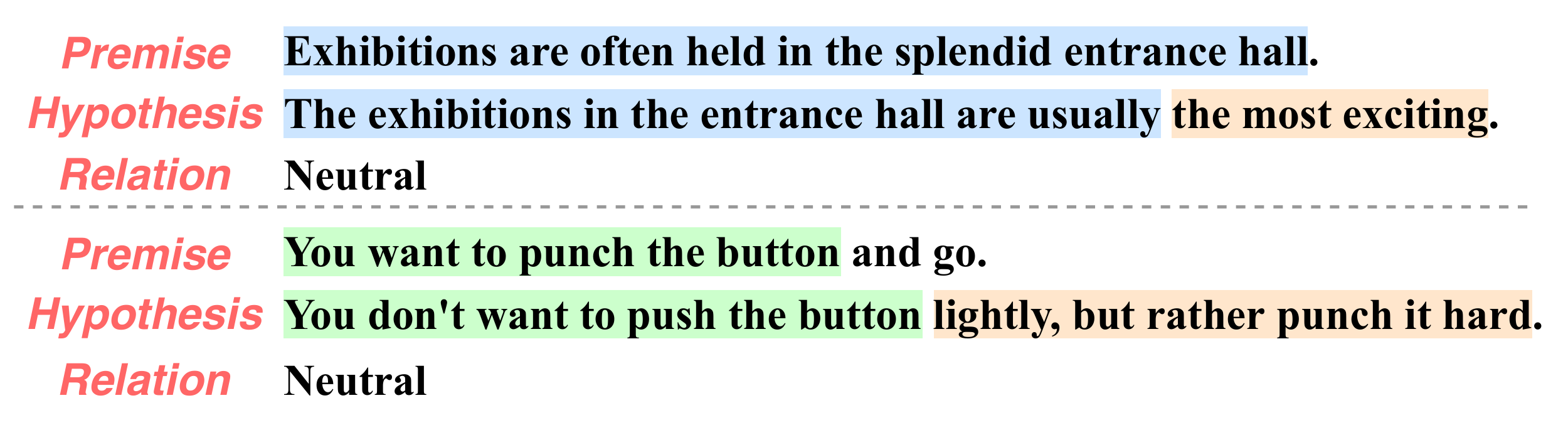}
    \caption{NLI instances from MultiNLI development sets. Text in blue/yellow/green are useful for the judgment of entailment/neutral/contradiction relations. For the first example, both the premise and hypothesis are about ``{\em the exhibition in the entrance hall\/}'', but the hypothesis points the ``{\em the most exiciting\/}'', which has no reference in the premise and imply the neutral relation. The second example is about that ``{\em you (don't) want to punch the button\/}'' which imply the contradiction relation. However, ``{\em punch it lightly or hard\/}'' in the hypothesis has no reference in the premise, so it is neutral.}
    \label{fig1}
\end{figure}

Natural language inference (NLI) aims to determine whether a natural-language hypothesis can be inferred from a given premise reasonably \cite{maccartney2009natural}, more concretely, the logic relationships including {\em entailment, neutral\/} and {\em contradiction\/}.

To determine the logic relationships effectively, various alignment mechanisms \cite{MacCartney2008APA,rocktaschel2015reasoning,guo2019gaussian} have been introduced into NLI models to model the interaction between the sentence pairs and achieve significant improvement in NLI. The achievement of the alignment mechanism is due to it being able to capture the aligned parts between the sentence pairs which are useful for the judgment of entailment and contradiction. 

However, we find that the aligned parts cannot benefit or even mislead the judgment of neutral (see Figure \ref{fig1}). We argue that almost all the parts in the sentence pairs are useful not limit to aligned parts, and the key point is their unequal contributions for the judgment of different relations. In fact, humans perform reasoning from multiple \textit{perspectives} associated with the three logic relationships, and focus on the different parts in each perspective to form a holistic perspective to eliminate bias for different relations.

Motivated by this, we propose the MPI, a novel NLI model that makes decision from the multiple perspectives associated with the three logic relationships. With MPI, all parts in the sentence pairs are captured unequally regarding different perspectives. To be specific, we formulate this procedure as {\em parts-to-wholes\/} assignment \cite{Sabour2017DynamicRB}, in which the tokens in the sentence pairs (parts) are assigned into the corresponding perspectives (wholes) they stand for. During this procedure, the MPI digs out the perspectives of different parts via routing-by-agreement \cite{Sabour2017DynamicRB} and takes all the perspectives into account to make the final decision.

Additionally, to ensure that the perspectives obtained by MPI are as expected, we introduce an explicit supervision on the representation of each perspective. Our explicit supervision does not only achieve better performance on the MPI, but also endows the obtained perspectives with high interpretability as well, which is absent in the previous routing-by-agreement-based studies of NLP.

Finally, experiments on SNLI and MultiNLI show that our MPI achieves significant improvement over the alignment based NLI models---cross alignment baseline and BERT, which also reveals the architecture-free capability of our MPI. Furthermore, we analyze the error rate by word overlap rate to demonstrate that our MPI relieves the misjudgment of entailment and neutral when the aligned parts are in a small amount, which is the error-prone area in NLI.

\begin{figure*}[t!]
\centering
    \centering
    \includegraphics[width=15cm]{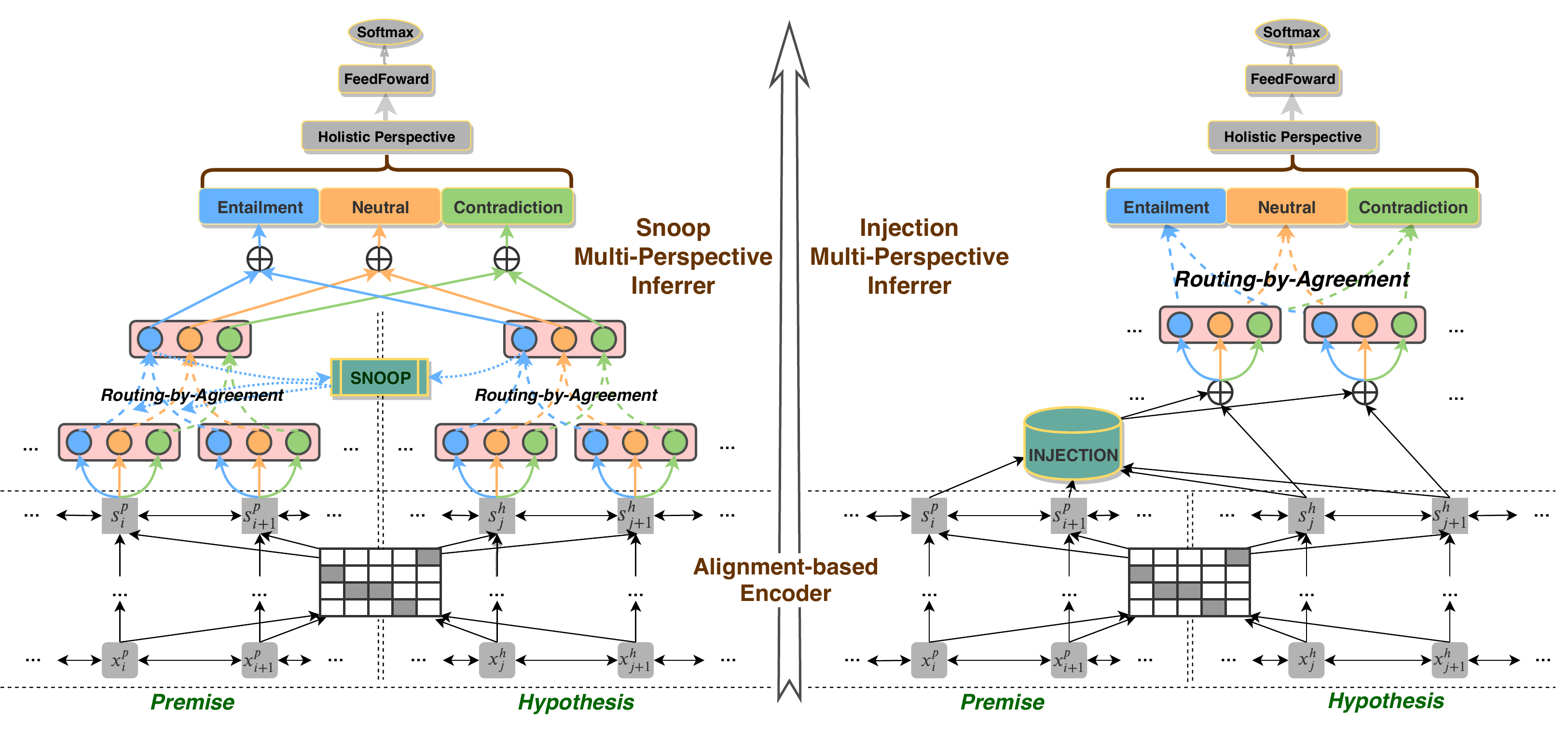}
    \caption{Multi-perspective inferrer (MPI) in alignment-based NLI model. The left one is the Snoop MPI and the right one is the Injection MPI. $\bigoplus$ denotes the concatenation operation. The alignment-based encoder can be instantiated as cross alignment encoder or BERT encoder.}
    \label{fig2}
\end{figure*}

\section{Background}

In this section, we summarize the alignment-based NLI models in a highly abstract way\footnote{See the appendix for the detail implementation.}. Based on this architecture, we introduce the proposed approach in Section \ref{mpi}.

\noindent{\bf Notation} We denote premise as $\mathbf{x}^P=(\mathbf{x}^P_1,\mathbf{x}^P_2,...,\mathbf{x}^P_m)$ and hypothesis as $\mathbf{x}^H=(\mathbf{x}^H_1,\mathbf{x}^H_2,...,\mathbf{x}^H_n)$, where $\mathbf{x}^P_i$ or $\mathbf{x}^H_j\in\mathbb{R}^K$ is a $K$-dimensional one-hot vector of the token. We use $\bar{\mathbf{x}}^{P}$ and $\bar{\mathbf{x}}^{H}$ to denote the word embedding (distribution representations) of $\mathbf{x}^{P}$ and $\mathbf{x}^{H}$.

\subsection{Alignment-based Encoder}

The alignment-based encoder in NLI aims to encode the context and model the interaction between the sentence pairs. The encoded representation of each token in the sentence pairs can be abstractly formulated as:
\begin{align}
(\mathbf{s}^{P}, \mathbf{s}^{H})=\text{Encoder}(\bar{\mathbf{x}}^{P},\bar{\mathbf{x}}^{H})
\end{align}
We divide the dominative alignment-based methods into two groups---cross alignment and BERT alignment.

\paragraph{Cross Alignment Encoder} In cross alignment encoder, premise and hypothesis are fed into the encoder separately. This type of encoder usually consists of one or several RNN/CNN/Transformer layers to obtain the context-aware representation of each token in the sentence pairs \cite{Chen2017EnhancedLF, Gong2018NaturalLI, guo2019gaussian}. :
\begin{align}
\bar{\mathbf{x}}^{P/H}=\text{RNN/CNN/Transformer}(\bar{\mathbf{x}}^{P/H}). 
\end{align}
To model the interaction between sentence pairs, various alignment mechanisms have been introduced among the encoder layers:
\begin{align}
(\tilde{\mathbf{x}}^P_i, \tilde{\mathbf{x}}^H_i)&=\text{Cross-Alignment}({\bar{\mathbf{x}}^P_i},\bar{\mathbf{x}}^H_j),
\end{align}
where Cross-Alignment can be dot-product attention \cite{Chen2017EnhancedLF}, convolutional interaction network \cite{Gong2018ConvolutionalIN}, factorization machines \cite{Tay2018CompareCA} and etc.
Finally, the representation of each token can be obtained as:
\begin{align}
\mathbf{s}^{P/H}_i=f_1(\bar{\mathbf{x}}^{P/H}_i, \tilde{\mathbf{x}}^{P/H}_i),
\end{align}
where $f_1$ can be various transformation like difference and element-wise product \cite{Mou2016NaturalLI}.

\paragraph{BERT Encoder}
Recently, BERT \cite{Devlin2018BERTPO} pre-trained on the large corpus achieves state-of-the-art performance on lots of NLP tasks. Different from the cross alignment encoder in which the sentence pair are fed separately, premise and hypothesis are concatenated as one sequence in BERT encoder. Then BERT inherently models the alignment among the tokens by self-attention in Transformer~\cite{Vaswani2017AttentionIA}:
\begin{align}
[\mathbf{s}^{P}; \mathbf{s}^{H}]=\text{BERT-Encoder}([\bar{\mathbf{x}}^{P};\bar{\mathbf{x}}^{H}]).
\end{align}

\subsection{Prediction}

Once we encode the sentence pairs, a prediction module is used to convert the obtained feature vectors $\mathbf{s}^{P}$ and $\mathbf{s}^{H}$ into a fixed length vector:
\begin{equation}
\mathbf{v}=f_2([\mathbf{s}^P;\mathbf{s}^H]),
\end{equation}
where $f_2$ is a multi-layer perceptron (MLP) with pooling. Then we predict the logic relationships of $(\mathbf{x}^P, \mathbf{x}^H)$ based on this fixed length vector:
\begin{align}
    p(l|\mathbf{x}^P, \mathbf{x}^H) = \mathrm{softmax}(E_l^\top \mathbf{v}),
    \label{eq:pred}
\end{align}
where $E_l$ is the label embedding for $l$ relationship.


\section{Multi-perspective Inferrer}\label{mpi}

In this section, we propose a novel inference method called Multi-Perspective Inferrer (MPI), which aims to pick out the perspectives of different parts in the sentence pairs and take all the perspectives into account to make the final decision.

Our inferrer is motivated by the observation: different parts in the sentence pairs imply different relationships (see Figure \ref{fig1}), and their significance is not equal for the judgment of different relationships. NLI models should exploit the multiple perspectives associated with the three logic relationships---entailment, neutral and contradiction. This procedure can be seen as a {\em parts-to-wholes\/} assignment \cite{Sabour2017DynamicRB}, where {\em parts\/} are tokens in the sentence pairs and {\em wholes\/} are the perspectives associated with the relations.

\newcite{Sabour2017DynamicRB} showed the capability of \textit{Capsule Networks} \cite{Hinton2011TransformingA} to solve the problems of assigning parts to wholes. A \textit{capsule} is a neuron representing one of the distinct properties (wholes) of the input features (parts) and can be determined by an iterative refinement procedure called \textit{dynamic routing} \cite{Sabour2017DynamicRB}. In the iterative routing procedure, the proportion of parts are assigned to wholes by the agreement between them. Obviously, we can employ this procedure in our MPI.

Figure \ref{fig2} depicts the overall architecture of the alignment-based model with MPI. Instead of pooling the extracted features to predict directly, first MPI picks out the existing perspectives from the extracted features of the sentence pairs via \textit{dynamic routing} and then takes all the perspectives into account to make the final prediction. The biggest challenge is how to employ the routing policy to solve the sentence-pair inputs rather than a single input. Therefore, we design two variants of MPI, the Snoop MPI and the Injection MPI, using different routing policies to pick out the perspectives in the sentence pairs.

\subsection{Modeling Perspectives via Routing-by-Agreement Policy}

First, we introduce the basic concept of the multi-perspective inferrer (MPI) via the routing-by-agreement policy, which determines the perspectives from premise and hypothesis respectively. We route the sentences pairs \textit{individually} using Algorithm \ref{alg:vdr} and then combine them by perspectives.

Take the hypothesis as an example. We define the extracted features $\mathbf{s}^H_i$ of token $i$ as low-level capsules and its perspectives representations $\mathbf{v}^H_l$ as high-level capsules, where $l \in L=\{\textsc{En,Ne,Con}\}$ is the perspective they stand for\footnote{{\em Orphan capsule} \cite{Sabour2017DynamicRB} are introduced to attract irrelevant assignment of noisy contents. For simplicity and correspondence with the perspectives we omitted it there.}. Before the routing process, each extracted features vector $\mathbf{s}^H_i \in \mathbb{R}^{d_{\text{low}}}$ is transformed into the ``vote vector'' $\mathbf{u}^H_{il}$:
\begin{equation}
\mathbf{u}^H_{il} = \mathbf{W}_l\mathbf{s}^H_i \in \mathbb{R}^{d_{\text{high}}},~~\forall l \in L,\label{vote}
\end{equation}
where $\mathbf{W}_l(\cdot)\in \mathbb{R}^{d_{\text{low}}\times d_{\text{high}}}$ is a trainable transformation matrix and shared among the tokens. Then MPI uses dynamic routing in Algorithm \ref{alg:vdr} to assign vote vectors $\mathbf{u}^H_i$ into the high-level capsules $\mathbf{v}^H_l$. The ``squash'' in Algorithm \ref{alg:vdr} is a non-linear activation function:
\begin{equation}
\text{squash}(\hat{\mathbf{v}}^H_l)=\frac{\|\hat{\mathbf{v}}^H_l\|^2}{1+\|\hat{\mathbf{v}}^H_l\|^2}\frac{\hat{\mathbf{v}}^H_l}{\|\hat{\mathbf{v}}^H_l\|},
\end{equation}
where $\hat{\mathbf{v}}^H_l$ is the inactivated high-level capsule representations by the weighted sum of the corresponding low-level ``vote vectors'':
\begin{equation}
\hat{\mathbf{v}}^H_l=\sum^n_i c^H_{il}\mathbf{u}^H_{il},
\end{equation}
where the assignment weight $c_{il}^H$ is computed by 
\begin{align}
\label{vbupdate}
b^H_{il}\leftarrow b^H_{il}+\mathbf{u}^H_{il}\cdot\mathbf{v}^H_l, \\
c_{il}^H = \mathrm{softmax}(b^H_{il}).
\end{align}

The high-level capsules of premise $\mathbf{v}^P_l$ can be obtained by the same procedure. Finally, the representation of each perspective is obtained by concatenating the high-level capsules of sentence pair:
\begin{equation}
\mathbf{v}_l=[\mathbf{v}^P_l;\mathbf{v}^H_l].\label{perspective}
\end{equation}

\subsubsection*{Snoop Multi-perspective Inferrer}
Although the final perspective representations obtained above are from the sentence pairs, it is supposed to generate the perspectives by the sentence pair mutually rather than individually. To remedy this, we introduce a variant of the routing policy called snoop routing (see the left of Figure \ref{fig2}). 

Take the hypothesis as an example. The snoop routing policy models its interaction with premise: when the low-level capsules of hypothesis update their routing weights, they will not only consider the agreement with their own high-level capsules $\mathbf{v}^H_l$, also snoop at the high-level capsules of premise $\mathbf{v}^{P}_l$\footnote{We also tried routing with the concatenated sentence pairs, but find that the sentence-level gap is much larger than the perspectives gap among the tokens. There are two main differences from snoop routing to concatenation routing: First, the high-level capsules are not fed with the low-level information from another sentence. Second, there is a trade-off between the own high-level capsules and the snooped high-level capsules during routing weights update.}. Formally, instead of updating routing weights $b^H_{il}$ in Equation \ref{vbupdate}, we update $b^H_{il}$ by:
\begin{equation}
b^H_{il}\leftarrow b^H_{il}+\mathbf{u}^H_{il}\cdot\mathbf{v}^H_l+\alpha\mathbf{u}^H_{il}\cdot\mathbf{v}^{P}_l,
\end{equation}
\begin{algorithm}
\caption{Dynamic Routing}
\label{alg:vdr}
\KwIn{vote vectors $\mathbf{u}_i$ in low-level, iterations $r$}
\KwOut{high-level $\mathbf{v}_l$}
$\textbf{procedure}$ ROUTING($\mathbf{u}_i$, $r$):\\
    $\forall i,l: b_{il}\leftarrow 0$\\
    \For{$r$ $\text{iterations}$}{
        for each $i$ in low-level, $l$ in high-level: $c_{il}\leftarrow \text{softmax}(b_{il})$\\
        for each $l$ in high-level: $\mathbf{v}_l\leftarrow\text{squash}(\sum_i c_{il}\mathbf{u}_{il})$\\
        for each $i$ in low-level, $l$ in high-level: $b_{il}\leftarrow b_{il}+\mathbf{u}_{il}\cdot\mathbf{v}_l$
    }
\textbf{return} $\mathbf{v}_l$
\end{algorithm}
\noindent where $\alpha$ is the snoop coefficient to model the similarity of the individual perspectives between the sentence pairs. Specifically, the snoop coefficient is generated by the high-level capsules of the sentence pairs:
\begin{equation}
\alpha=\sigma(\text{Linear}([\mathbf{v}^H_l;\mathbf{v}^P_l])).
\end{equation}

\subsubsection*{Injection Multi-perspective Inferrer}
In contrast to the Snoop MPI, which uses a routing policy on both premise and hypothesis to obtain the final perspectives, another form of MPI called Injection Multi-perspective Inferrer (IMPI) determines the perspectives mainly focusing on hypothesis (see the right of Figure \ref{fig2}), which is inspired by the intrinsic asymmetry of NLI \cite{MacCartney2008APA} and the fact of the perspectives mainly expressed by the hypothesis.

However, the information of the premise will leak while only routing the hypothesis to form the final perspectives. So the key issue of IMPI is to inject the information of premise into the hypothesis. To do so, we resort to bilinear attention as the injection method:
\begin{align}
&~~~~~~~~~~~~~~~~~~~~~\alpha_{ij}=\mathbf{s}^P_i\mathbf{W}\mathbf{s}^H_j, \\
\tilde{\mathbf{s}}^H_j&=\sum_{i=1}^{m}\text{softmax}_i(\alpha_{ij})\bar{\mathbf{s}}^P_i,\forall j\in[1,...,n].
\end{align}

Finally, we use $\hat{\mathbf{s}}^H_i=[\mathbf{s}^H_i;\tilde{\mathbf{s}}^H_i]$ instead of $\mathbf{s}^H_i$ in Equation \ref{vote} to get the vote vectors and use Algorithm \ref{alg:vdr} to get the final perspective $\mathbf{v}_l=\mathbf{v}^P_l$.

\subsection{Prediction from Holistic Perspective}
After computing the representations of each perspective, we can obtain a holistic perspective representation by:
\begin{align}
  \mathbf{v} =  f_1([\mathbf{v}_\textsc{EN},\mathbf{v}_\textsc{NE},\mathbf{v}_\textsc{CON}]),
\end{align}
where $f_1$ is a two-layer feed-forward network. Then we predict the logic relationships of $(\mathbf{x}^P, \mathbf{x}^H)$ based on this holistic perspective.
\paragraph{Training} We use multi-class cross-entropy loss to optimize our base model by minimizing the negative log-likelihood for each pair of $(\mathbf{x}^P, \mathbf{x}^H)$ in the dataset of i.i.d observations:
\begin{align}
    \ell = - \log p(l|\mathbf{x}^P, \mathbf{x}^H).
    \label{eq:obj}
\end{align}

\subsection{Learning Perspectives as Expected}\label{loss}
To ensure that the obtained perspectives actually represent the associated relations, we design an auxiliary loss for the representations of perspectives. Formally, $\mathbf{v}_l$ is required to be predictive of $l$ rather than the other relations, and the corresponding probability is computed by:
\begin{align}
    \Tilde{\mathbf{v}}_l &= f_2(\mathbf{v}_l), \\
    p(l|\mathbf{v}_l) &= \mathrm{softmax}(E_l^\top \Tilde{\mathbf{v}}_l),
\end{align}
where $f_2$ is a two-layer feed-forward network.
Consider that one perspective, say $l$, does not exist in some cases, then the corresponding prediction $p(l|\mathbf{v}_l)$ will be non-informative. Thus, we resort to the \textit{weighted cross-entropy} loss by taking the existence of each perspective into account, and maximize $p(l|\mathbf{v}_l)$ for each perspective representation:
\begin{align}
    \ell_{\mathrm{ce}}(\mathbf{v}) = -\sum_{l \in L} \|\mathbf{v}_l\| \cdot \log p(l|\mathbf{v}_l),
    \label{eq:aux}
\end{align}
where $\|\mathbf{v}_l\|$ is the norm length of $\mathbf{v}_l$, representing the existence of the perspective $l$ due to the property of capusle network~\cite{Sabour2017DynamicRB}. We further regularize the norm length of each capsule by \textit{marginal loss} to ensure the least existence of each perspective:
\begin{align}
    \ell_{\mathrm{aux}} = \ell_{\mathrm{ce}}(\mathbf{v})  + \ell_{\mathrm{margin}}(\|\mathbf{v}\|).
\end{align}
The final training objective is updated from Equation \ref{eq:obj}  to $\ell^{*} = \ell + \beta \cdot \ell_{\mathrm{aux}}$, where $\beta$ is a hyper-parameter.

\section{Experiments}

\subsection{Datasets}

We use the Stanford Natural Language Inference (SNLI) dataset \cite{Bowman2015ALA} and the Multi-Genre Natural Language Inference (MultiNLI) dataset \cite{Williams2018ABC} to make quantitative evaluation of our MPIs. These two datasets both focus on three concrete logic relations between the given premise and hypothesis: {\em entailment, neutral\/} and {\em contradiction\/}.

\noindent{\bf SNLI} The SNLI corpus consists of 570k sentence pairs. The premise data is drawn from the captions of the Flickr30k corpus, and the hypothesis data is manually composed.
We discard the ``$-$'' annotated relationship (lack of human annotation) as with previous work. 
The training/development/test datasets consist of 549,367/9,842/9,824 pairs of sentence. 

\noindent{\bf MultiNLI} The MultiNLI corpus consists of 433k sentence pairs. These sentence pairs contain nine genres, which compose the concept of multi-genres. In MultiNLI, only half of genres appear in the training set while the rest are not, creating matched (in-domain) and mismatched (cross-domain) development/test sets. We use the official training/development sets to select our best models and test on Kaggle.com\footnote{\url{https://www.kaggle.com/c/multinli-matched-open-evaluation/leaderboard}, and \url{https://www.kaggle.com/c/multinli-mismatched-open-evaluation/leaderboard}}.

We use the accuracy to evaluate the performance
of our multi-perspective inferrers (MPIs) and other
models on SNLI and MultiNLI.

\subsection{Overall Results}

Table \ref{tb1:nlioverall} shows the performance of the state-of-the-art models and the proposed MPIs. Our approach achieves the best performance on both SNLI (BERT+IMPI) and MultiNLI (BERT+IMPI) test datasets compared to the previous studies. In detail, our MPIs achieve 1.2 percent improvement over cross alignment baseline, and 0.7 percent improvement over BERT baseline. The significant improvement on two NLI datasets demonstrates that our motivation of inferring relationships from multiple perspectives does make sense.

\begin{table}[t]
\begin{tabular*}{\hsize}{l@{}c@{}c}
\toprule
\bf Models         & \bf ~SNLI~ & \bf MultiNLI \\ 
\midrule
DecompAtt+IntraAtt~& 86.8 & -/- \\
~\cite{parikh2016decomposable} \\
BiMPM~\cite{Wang2017BilateralMM} & 86.9 & -/- \\
{\it ESIM} \cite{Chen2017EnhancedLF} & 88.0 & 76.8/75.8 \\
CIN~\cite{Gong2018ConvolutionalIN} & 88.0 & 77.0/77.6 \\
DIIN~\cite{Gong2018NaturalLI} & 88.0 & 78.8/77.8 \\
MwAN~\cite{Tan2018MultiwayAN} & 88.3 & 78.5/77.7 \\
CAFE~\cite{Tay2018CompareCA} & 88.5 & 78.7/77.9 \\
KIM~\cite{chen2017neural} & 88.6 & 77.2/76.4 \\
SAN~\cite{Liu2018StochasticAN} & 88.7 & 79.3/78.7 \\
DRCN~\cite{Kim2019SemanticSM} & 88.9 & 79.1/78.4 \\
Gaussian Transformer      & \textbf{89.2} & 80.0/79.4 \\
~\cite{guo2019gaussian} \\
$\textit{BERT}_\text{base}$~\cite{Devlin2018BERTPO} & -    & \textbf{84.6}/\textbf{83.4} \\
\hline Cross Alignment baseline     & 87.9 & 77.7/76.8 \\
+ Vanilla MPI         & 88.4 & 78.6/77.7 \\
+ SMPI                & 88.6 & 78.7/77.9 \\
+ IMPI                & \textbf{88.7} & \textbf{78.9/78.1} \\
\hline 
BERT baseline ($v_{\text{test}}=1.0\times$)     & 90.2    & 84.5/83.3 \\
+ Vanilla MPI ($v_{\text{test}}=0.98\times$)         & 90.3    & 84.6/83.5 \\
+ SMPI  ($v_{\text{test}}=0.98\times$)               & 90.7    & 85.0/83.8 \\
+ IMPI  ($v_{\text{test}}=0.99\times$)             & \textbf{90.8}    & \textbf{85.2}/\textbf{84.1} \\
\toprule
\end{tabular*}
\caption{Performance on SNLI and MultiNLI (matched/ mismatched) test datasets. Our MPIs models are implemented on both cross alignment baseline and BERT. $v_{\text{test}}$ is the test speed compared to the baseline.}
\label{tb1:nlioverall}
\end{table}
\paragraph{Architecture-free Capability of MPIs} 
Theoretically, our MPIs could be architecture-free, since the MPIs' inputs are the extracted features of each token in the premise and hypothesis (see Figure \ref{fig2}) no matter what kind of the encoder we use. As shown in Table \ref{tb1:nlioverall}, the significant improvement on both cross alignment baseline and BERT demonstrates the architecture-free capability of our MPIs. We suggest that modeling multiple perspectives is important and useful for NLI, and the proposed solution could inspire future research in this filed.

\paragraph{SMPI v.s IMPI}
As shown in Table \ref{tb1:nlioverall}, both variants of the MPI get better performance than the vanilla MPI, which demonstrates that the perspectives should be generated by the sentence pairs together and the snoop and injection mechanisms are powerful. 
Additionally, we find that the IMPI always gets better performance than the SMPI on both SNLI and MultiNLI, which may relate to the intrinsic asymmetry in NLI \cite{MacCartney2008APA} and the fact of the perspectives mainly expressed by the hypothesis. \newcite{Poliak2018HypothesisOB} pointed out that it is related to the irregularities of the datasets. However, we argue that the sentence pairs in NLI are actually intrinsically asymmetrical and we should deal with the premise and hypothesis differently.

\paragraph{Efficiency}
As shown in the last group of Table \ref{tb1:nlioverall}, we perform the efficiency test on BERT baseline and our MPIs. The test speed of Vanilla MPI and SMPI is 0.98 times compared to the BERT baseline while the test speed of IMPI is 0.99 times compared to the BERT baseline. Our MPIs improve the baseline system with very mild speed degradation, which demonstrates a good efficiency of MPIs that could be applied to arbitrary architecture with less hurt.


\subsection{Analysis and Discussion}
To verify our motivation and investigate if our MPIs did relieve the error phenomenon in our observation, we take the cross alignment baseline as an example to conduct in-depth analysis and discussion on word overlap rate and auxiliary loss.

\paragraph{Word Overlap Rate} Word overlap between the sentence pairs is a special case of alignment. One of our observations in the base model is that: if the aligned parts between sentence pairs are in a small amount, the misjudgments of entailment and neutral will be serious. To investigate whether our MPIs relieve this phenomenon, we perform the analysis of the error rate by word overlap rate on MultiNLI development matched set.

Formally, given a sentence pair of premise $\mathbf{x}^P=\langle x^P_1,x^P_2,...,x^P_m \rangle$ and hypothesis $\mathbf{x}^H= \langle x^H_1,x^H_2,...,x^H_n \rangle$, the word overlap rate $R_{\text{wo}}$ from hypothesis to premise can be defined as:
\begin{equation}
R_{\text{wo}}=\frac{\sum^n_{j=1} \mathds{1}(x^P_j \in \mathbf{x}^H)}{m}.
\end{equation}
To be specific, we divide MultiNLI development matched set into six groups by word overlap rates: [0,0.2), [0.2,0.4), [0.4,0.6), [0.6,0.8), [0.8,1.0) and 1.0, and Table \ref{woamount} shows the statistics.

Then we compute the error rates of entailment and neutral individually\footnote{See appendix for total and contradiction.}. 
Figure \ref{fig:hyper2} shows the error rate of entailment class. Both two MPIs reduce the error rates of low word overlap rates, i.e., [0,0.2) and [0.2, 0.4), significantly. This is because the judgment of entailment is related to the alignment, and the base model cannot utilize adequate aligned parts to make the prediction of entailment. After applying MPI, the aligned parts can be captured by entailment perspective although it is in a small amount. Then the classifier can focus more on the entailment perspective so the error rate of these interval decreases significantly.
Figure \ref{fig:hyper3} shows the error rate of the neutral class. Our MPIs reduce the error rates of the low word overlap rate intervals, i.e., [0,0.2) and [0.2, 0.4), the same as with the entailment class. Consider both entailment class and neutral class, our MPIs obviously relieve the phenomenon: the small amount aligned parts can cause the misjudgment between entailment relations and neutral relations.

\begin{table}[t!]
\centering
\footnotesize
\begin{tabular}{|c|c|c|c|}
\hline \bf Relation & \bf [0,0.2) & \bf [0.2,0.4) & \bf [0.4,0.6) \\
\hline
\textsc{En} & 458 & 1331 & 986 \\
\textsc{Ne} & 982 & 1254 & 495\\
\textsc{Con} & 845 & 1368 & 651  \\
\hline
Total & 2285 & 3953 & 2132 \\
\hline
\bf Relation & \bf [0.6,0.8) & \bf [0.8,1.0) & \bf 1.0 \\
\hline
\textsc{En} & 513 & 147 & 44 \\
\textsc{Ne} & 253 & 68 & 71 \\
\textsc{Con} & 284 & 54 & 11 \\
\hline
Total & 1050 & 269 & 126 \\
\hline
\end{tabular}
\caption{Statistics of MultiNLI development matched set divided by the word overlap rates into six groups: [0,0.2), [0.2,0.4), [0.4,0.6), [0.6,0.8), [0.8,1.0) and 1.0.}\smallskip
\label{woamount}
\end{table}

\begin{figure}[t] 
  \centering
  \subfigure[Entailment.] {
    \label{fig:hyper2}
    \includegraphics[width=0.45\linewidth]{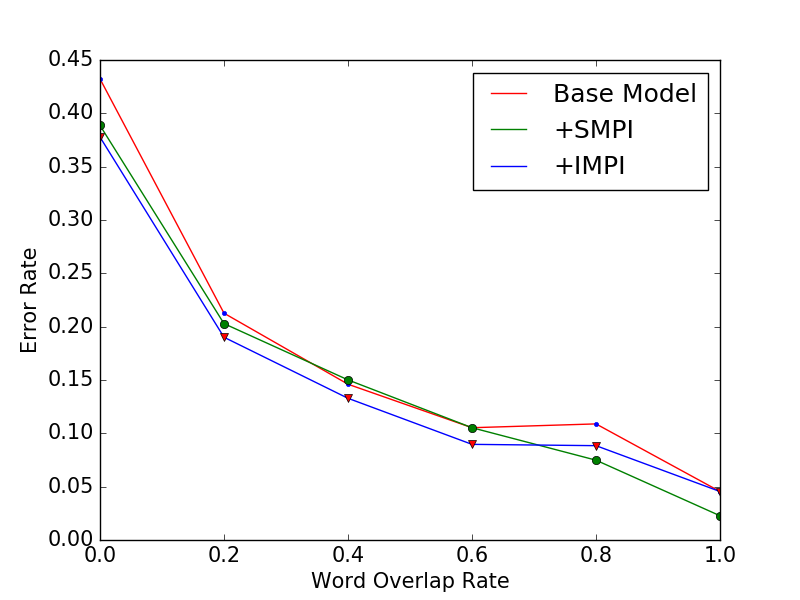}
  }
  \hspace{0.01\linewidth}
  \subfigure[Neutral.] {
    \label{fig:hyper3}
    \includegraphics[width=0.45\linewidth]{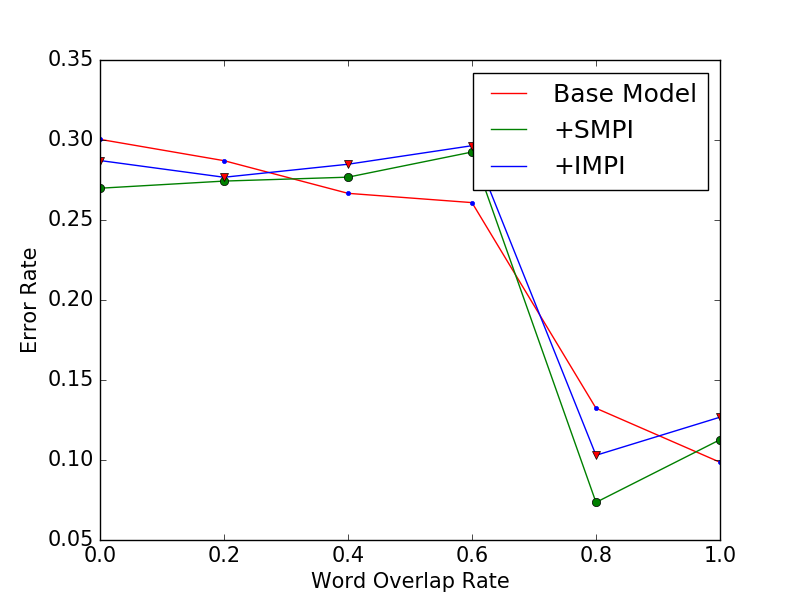}
  }
  \caption{Error rate of entailment and neutral by word overlap rate. Base Model is the cross alignment baseline. Our MPIs relieve the error rate in the intervals with low word overlap rate [0, 0.4) of entailment and neutral significantly, which is the most error-prone area in NLI.}
  \label{fig:hyper}
\end{figure}

\noindent{\textbf{\textit{Disaster Area in NLI: lower word overlap rates.}}} 
We find a similar phenomenon concerning the error rate and the amount of entailment and neutral: the error is much more severe in the lower word overlap rate intervals (e.g., [0,0.2) and [0.2, 0.4)) than the higher rate intervals. If the aligned parts are in a small amount, the strength of alignment-based models will be weakened. It is the most error-prone area in NLI. We believe that NLI models need not only focus on capturing the aligned parts, but also take advantage of the unaligned parts. To summarize, NLI models should take the problems into account in multiple perspectives and solve them from a holistic perspective.

\paragraph{Auxiliary Loss}
Our auxiliary loss is an explicit supervision on the representation of each perspective to ensure the perspectives obtained by our MPIs are expected. In this section, we demonstrate that our auxiliary loss improves both performance and an interpretability of predictions.

First, we compute the accuracy of predictions by perspective representations $\mathbf{v}^n_l$ according to Equation \ref{auxacc} and obtain 100\% accuracy on MultiNLI development matched set. It reveals that our auxiliary loss ensures that the representation of each perspective is associated with the corresponding relationship.
\begin{align}
    acc = \frac{1}{|L|\cdot N} \sum_{n = 1}^N \sum_{l \in L} \mathbf{1}(p(l|\mathbf{v}^n_l) == l).\label{auxacc}
\end{align}
\noindent{\textit{\textbf{The auxiliary loss improves the performance of the MPIs.}}}
Table \ref{ablations} shows the ablation studies of auxiliary loss on MultiNLI development sets. Results in the second line are the performance of two MPIs with the introduced supervision loss, which achieve the best accuracy. After removing the auxiliary loss, we find the performance of both two MPIs degrade to 78.7/78.4 and 78.6/78.4, which means the guided loss ensures the representation of each perspective.

\begin{table}[t!]
\centering
\footnotesize
\begin{tabular}{|l|c|c|}
\hline \bf Auxiliary Loss & \bf SMPI & \bf IMPI \\
\hline
w/o $\ell_{\rm aux}$ & 78.7/78.4 & 78.6/78.4 \\
\hline
w/ ~$\ell_{\rm aux}$ (i.e., $\ell_{\rm ce}$+$\ell_{\rm margin}$) &\bf 78.8/78.6 & \bf
79.1/78.9 \\
~~~~~~~~ - $\|\mathbf{v}_l\|$ & 78.7/78.7 & 78.3/78.8 \\
\hline
only w/ ~$\ell_{\rm ce}$ & 78.5/78.6 & 78.6/78.4 \\
~~~~~~~~ - $\|\mathbf{v}_l\|$ & 78.0/78.6 & 78.6/78.3 \\
\hline
only w/ ~$\ell_{\rm margin}$ & 78.6/78.2 & 78.4/77.8 \\
\hline
\end{tabular}
\caption{\label{ablations} Ablation studies of auxiliary loss on MultiNLI development sets.}\smallskip
\end{table}

\begin{figure*}[t]
    \centering
    \includegraphics[width=14.5cm]{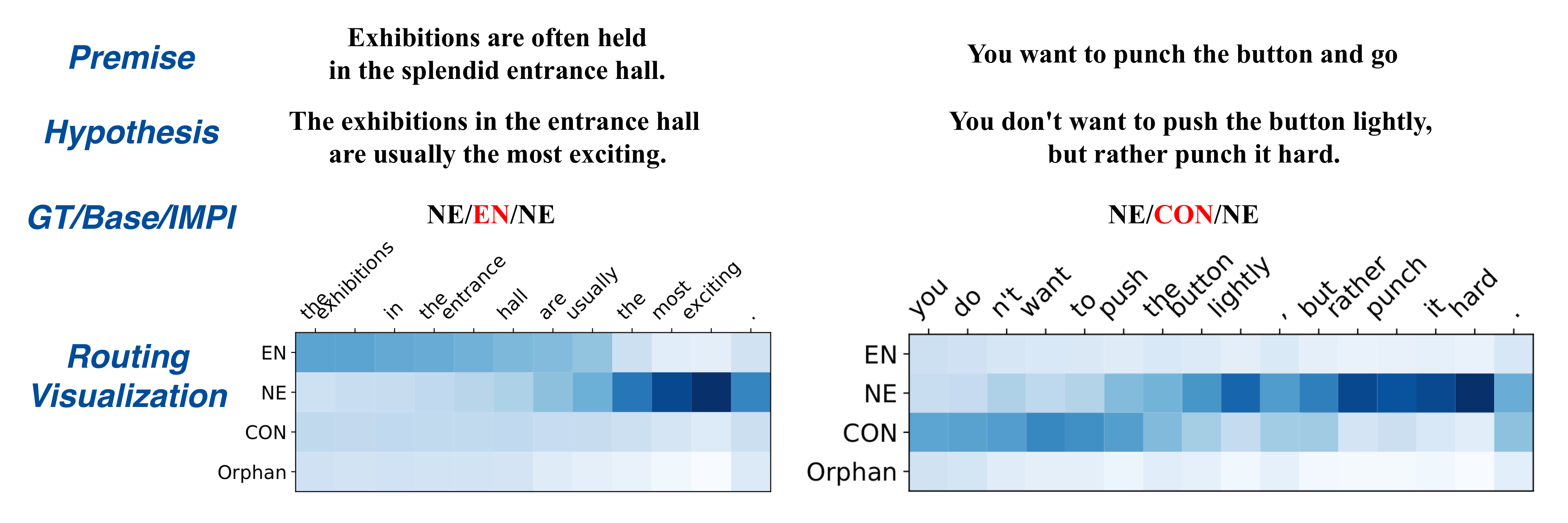}
    \caption{Visualization of routing weights $c_{il}$ in IMPI. Darker color corresponds to higher routing weights. GT represents the ground truth. Our base model gives the wrong predictions of both two samples. IMPI corrects the wrongs by taking multiple perspectives into account. Also, the routing weights of IMPI are highly interpretable. For the left example, IMPI routs the ``The exhibitions in the entrance hall'' to entailment perspective, which can be entailed by the premise actually. Additionally, IMPI routs the ``The the most exciting'' to neutral perspective since there is no evidence in the premise. For the right example, IMPI routs the ``you don't want to push the butthon'' to contradiction perspective, which conflicts with the premise actually. However, IMPI finds the more important parts ``push the button lightly, but rather punch it hard.'' in neutral perspective.}
    \label{fig5}
\end{figure*}

\noindent{\textit{\textbf{Visualization on Routing Assignment Weights.}}}
Another contribution of our auxiliary loss on each perspective is that the predictions of MPIs are highly interpretable. Figure \ref{fig5} contains two instances from MultiNLI development matched set. Since different parts of the sentences imply the different perspectives, our base model gets the incorrect predictions. However, capturing the different parts by multiple perspectives and making the decision from the holistic perspective accounts for our IMPI's correct predictions.\footnote{See other MPIs' routing visualization in the appendix.}

\section{Related Work}

Natural language inference is a long-standing problem in NLP research. Early studies use small datasets while leveraging lexical and syntactic features for NLI \cite{maccartney2009natural}. With the development of deep learning and the releases of large-scale annotated datasets, e.g. the SNLI \cite{Bowman2015ALA} and the MultiNLI \cite{Williams2018ABC}, researchers have made big progress on neural network based NLI models.

\newcite{MacCartney2008APA} introduced the traditional alignment mechanism into NLI and \newcite{rocktaschel2015reasoning} first proposed an attention-based neural model for NLI. After that, neural alignment-based models have been developed rapidly in NLI \cite{Wang2016LearningNL,Chen2017EnhancedLF,Ghaeini2018DRBiLSTMDR,Gong2018NaturalLI,Tay2018CompareCA,Gong2018ConvolutionalIN,Tan2018MultiwayAN,Kim2019SemanticSM,guo2019gaussian}. 
\newcite{parikh2016decomposable} introduced to decompose the different parts of sentences by decomposable attention. \newcite{Wang2017BilateralMM} proposed a multi-perspective matching model to capture the aligned parts by four matching operations. Both of them focus on capturing the aligned parts in multiple perspectives. However, our MPIs focus on how to utilize the obtained information (both aligned and unaligned parts) from multiple perspectives associated with the relationships. Motivated by this, we proposed a multi-perspective inferrer for existing alignment-based model inspired by \textit{parts-to-wholes} \cite{Sabour2017DynamicRB}.

\newcite{Sabour2017DynamicRB} introduced the capsule networks to learn parts-to-wholes relations automatically by dynamic routing, which has been applied to many NLP tasks recently such as sentiment analysis \cite{Wang2018SentimentAB,Wang2019AspectlevelSA}, relation extraction \cite{Zhang2018AttentionBasedCN,Zhang2019MultilabeledRE}, text classification \cite{Yang2018InvestigatingCN} and machine translation \cite{Dou2019DynamicLA,Zheng2019DynamicPA}. \newcite{Gong2018InformationAV} regard dynamic routing as an information aggregation mechanism by concatenating all capsules representations and feeding them into a classifier without supervision. There are two main differences from our MPIs: First, we design the enhanced routing policies---snoop and injection, based on the understanding of the property of perspectives and the intrinsic asymmetry of NLI. Second, we introduce an auxiliary loss as supervision on each perspective, which guarantees that the meaning of perspectives and these predictions are highly interpretable.

\section{Conclusion}

In this paper, we proposed a new inference method called multi-perspective inferrer (MPI) for NLI. The MPI extracts multiple perspectives associated with the relationships from the different parts of the sentence pairs and takes all the perspectives into account to make the final decision. Furthermore, we introduced an explicit supervision on each perspective which ensures the predictions are highly interpretable. Experiments on SNLI and MultiNLI demonstrate that our MPI achieves substantial improvement and especially relieves the misjudgment between entailment and neutral relations when there is a small amount of aligned parts, which is the most error-prone in NLI.
In future work, we hope to make the routing-by-agreement more suitable for NLI and other sentence pair tasks such as semantic matching and machine comprehension.


\bibliography{mpi4nli.bbl}
\bibliographystyle{acl_natbib}

\clearpage

\appendix

\section{Our Baselines for NLI}\label{abm4nli}

In this section, we introduce two baselines using different interactive encoders---cross alignment encoder and BERT encoder in detail, which are used as our baselines in the experiments.

\noindent{\bf Notation} We denote premise as $\mathbf{x}^P=(\mathbf{x}^P_1,\mathbf{x}^P_2,...,\mathbf{x}^P_m)$ and hypothesis as $\mathbf{x}^H=(\mathbf{x}^H_1,\mathbf{x}^H_2,...,\mathbf{x}^H_n)$, where $\mathbf{x}^P_i$ or $\mathbf{x}^H_j\in\mathbb{R}^K$ is a $K$-dimensional vector of the token.

\subsection{Cross Alignment Baseline}

In cross alignment baseline, premise and hypothesis are fed into the model separately. We use RNNs to learn the context-aware representation of each token in the sentence pairs and dot-product attention to capture the alignments between the premise and hypothesis.

\paragraph{Embedding}\label{emb}
We use {\em GloVe\/} as the pretrained word embeddings for the ESIM, as well as widely-used one-hot external linguistic features such as lexical, syntactical and part-of-speech tagging to enrich the representation of the inputs as with \newcite{Gong2018ConvolutionalIN,Gong2018NaturalLI}. All of these embeddings are fixed during training. They are together fed into a one-layer feed-forward network to construct the final input representations:
\begin{equation}
\bar{\mathbf{x}}=g(\mathbf{x})\cdot f(\mathbf{x})+(1-g(\mathbf{x}))\cdot \mathbf{x},\label{ffn}
\end{equation}
where $g(\cdot)$ is a fusion gate parameterized by MLP with {\em sigmoid\/} and $f(\cdot)$ is another MLP with {\em ReLU\/}.

\paragraph{Cross Alignment Encoding} 
First, we use an identical BiLSTM \cite{Hochreiter1997LongSM} to encode premise and hypothesis to get the context-aware representation of each token:
\begin{align}
\bar{\mathbf{x}}^P_i&=\text{BiLSTM}_1(\mathbf{x}^P, i),\forall i\in[1,...,m], \\
\bar{\mathbf{x}}^H_j&=\text{BiLSTM}_1(\mathbf{x}^H, j),\forall j\in[1,...,n].
\end{align}
Then, to model the local inference between sentence pairs, we use the dot-product attention to obtain the interaction representations:
\begin{align}
&~~~~~~~~~~~~~~~~~~~~~e_{ij}={\bar{\mathbf{x}}^P_i}{}^\top\bar{\mathbf{x}}^H_j, \\
\tilde{\mathbf{x}}^P_i&=\sum_{j=1}^{n}\text{softmax}_j(e_{ij})\bar{\mathbf{x}}^H_j,\forall i\in[1,...,m],\\
\tilde{\mathbf{x}}^H_j&=\sum_{i=1}^{m}\text{softmax}_i(e_{ij})\bar{\mathbf{x}}^P_i,\forall j\in[1,...,n].
\end{align}
After obtaining the encoded representations and interaction representations, we use the following enhancement by difference and element-wise product to take advantage of them \cite{Mou2016NaturalLI}:
\begin{align}
\mathbf{m}^P_i&=[\bar{\mathbf{x}}^P_i;\tilde{\mathbf{x}}^P_i;\bar{\mathbf{x}}^P_i-\tilde{\mathbf{x}}^P_i;\bar{\mathbf{x}}^P_i\odot\tilde{\mathbf{x}}^P_i],\\
\mathbf{m}^H_j&=[\bar{\mathbf{x}}^H_j;\tilde{\mathbf{x}}^H_j;\bar{\mathbf{x}}^H_j-\tilde{\mathbf{x}}^H_j;\bar{\mathbf{x}}^H_j\odot\tilde{\mathbf{x}}^H_j].
\end{align}
A one-layer feed-forward network to decrease the dimension of enhancement representations:
\begin{align}
\bar{\mathbf{m}}^P_i=f(\mathbf{m}^P_i),~~~~\bar{\mathbf{m}}^H_j=f(\mathbf{m}^H_j).
\end{align}
Finally, another BiLSTM is used to compose the enhancement representations with their context:
\begin{align}
\mathbf{s}^P_i&=\text{BiLSTM}_2(\bar{\mathbf{m}}^P, i),\forall i\in[1,...,m], \\
\mathbf{s}^H_j&=\text{BiLSTM}_2(\bar{\mathbf{m}}^H, j),\forall j\in[1,...,n].
\end{align}


\paragraph{Prediction} First, we use max pooling and average pooling to convert the obtained feature vectors into a fixed length vector: 
\begin{equation}
\mathbf{v}=f([\mathbf{s}^P_{\max};\mathbf{s}^P_{\text{avg}};\mathbf{s}^H_{\max};\mathbf{s}^H_{\text{avg}}]),
\label{eq:pooling}
\end{equation}
where $f$ is a multi-layer perceptron (MLP). Then we predict the logic relationships of $(\mathbf{x}^P, \mathbf{x}^H)$ based on this fixed length vector:
\begin{align}
    p(l|\mathbf{x}^P, \mathbf{x}^H) = \mathrm{softmax}(E_l^\top \mathbf{v}),
\label{eq:pred_appendix}
\end{align}
where $E_l$ is the label embedding for $l$ relationship.

\subsection{BERT Baseline}

In BERT~\cite{Devlin2018BERTPO} baseline, premise and hypothesis are concatenated as one sequence to feed into the model instead of feeding separately. We use the Google BERT pretrained on the large corpus and fine-tunes it on the NLI datasets.

\paragraph{Embedding}
Since the premise and hypothesis are concatenated as one sequence to feed into the model, a special token [SEP] is inserted between them to differentiate the two sentences. Additionally, another special token [CLS] is inserted as the first token, which will be used to make the final prediction in the original BERT implementation:
\begin{align}
    \bar{\mathbf{x}}&=[[\text{CLS}];\bar{\mathbf{x}}^P;[\text{SEP}];\bar{\mathbf{x}}^H;[\text{SEP}]]. 
\end{align}

\paragraph{BERT Encoding}
The self-attention mechanism in Transformer~\cite{Vaswani2017AttentionIA} performs the alignment role among the tokens in the concatenated premise and hypothesis. Finally, we obtain the extracted features of each token:
\begin{align}
\mathbf{s}&=\text{BERT-Encoder}(\bar{\mathbf{x}}).
\end{align}

\paragraph{Prediction}
In the original BERT, the extracted features of special token [CLS] are used to make the final prediction directly. In our BERT-based model, we use the extracted features of each token to conduct average pooling instead\footnote{We found the performance of these two methods is close in the preliminary experiments.}:
\begin{align}
p(l|\mathbf{x}^P, \mathbf{x}^H) = \mathrm{softmax}(E_l^\top \cdot f(\mathbf{s}_\text{avg})).
\end{align}


\section{Implementation Details}

We implement our models using PyTorch and train them on Nvidia 1080Ti. We use the Adam optimizer \cite{Kingma2015AdamAM} with an initial learning rate of 0.0002. If the loss on development sets doesn't decrease in three epochs, the learning rate will be decayed to half. The minimum learning rate is $10^{-5}$. The dropout rate of 0.4 is applied before each feed-forward layer and recurrent layer. L2 regularization is set to $10^{-5}$. The batch size is set to 32. The hidden size is set to 300. All parameters are initialized with xavier initialization \cite{Glorot2010UnderstandingTD}. Word embeddings are preloaded with 300d GloVe embeddings \cite{Pennington2014GloveGV} and fixed during training. The iteration of routing is tuned amongst \{1, 2, 3, 4, 5\}. The $m^+$ and $m^-$ of marginal loss is tuned amongst \{0.8, 0.9\} and \{0.4, 0.5\}.

For the BERT baseline, we use the official uncased BERT-Base (12-layer, 768-hidden, 12-heads, 110M parameters) and fine-tune it with our MPIs according to the preliminary experiments on the cross alignment baseline.

\section{Linguistic Error Analysis}

We perform the linguistic error analysis using the supplementary annotations provided by the MultiNLI dataset. We compare the MPIs against the outputs of the our base model across 13 categories of linguistic phenomena. Table \ref{lingm} and Table \ref{lingmm} show the results. Both two MPIs outperform our base model on overall accuracy on the matched and mismatched sets. The word overlap category in the supplementary annotations is whether both sentences share more than 70\% of their tokens, containing 37 instances. Their statistical method of word overlap is much different from ours: the word overlap rate $R_{\text{wo}}$ is calculated by the amount of shared tokens and the amount of hypothesis, which is related to the intrinsically asymmetric in NLI \cite{MacCartney2008APA} and the fact that the perspectives mainly is expressed by the hypothesis.

\begin{table}[t!]
\centering
\begin{tabular}{|l|c|c|c|}
\hline \bf Category & \bf Baseline & \bf SMPI & \bf IMPI \\
\hline
Conditional & 69.6 & \textbf{73.9} & 56.5 \\
Word overlap & \textbf{92.9} & 89.3 & 89.3 \\
Negation & \textbf{82.2} & 77.5 & 79.8 \\
Antonym & \textbf{76.5} & 64.7 & \textbf{76.5} \\
Long Sentence & \textbf{78.8} & 75.8 & \textbf{78.8} \\
Tense Difference & 82.4 & \textbf{84.3} & \textbf{84.3} \\
Active/Passive & \textbf{100.0} & 93.3 & \textbf{100.0} \\
Paraphrase & \textbf{96.0} & 92.0 & 92.0 \\
Quantity/Time & \textbf{66.7} & \textbf{66.7} & 60.0 \\
Coreference & \textbf{73.3} & \textbf{73.3} & \textbf{73.3} \\
Quantifier & \textbf{80.8} & 76.8 & 80.0 \\
Modal & \textbf{80.6} & 79.2 & \textbf{80.6} \\
Belief & 77.3 & 72.7 & \textbf{78.8} \\
\hline
Overall & 77.8 & 78.7 & \textbf{79.1} \\
\hline
\end{tabular}
\caption{Linguistic Error Analysis on MultiNLI development matched sets. Baseline is the cross alignment baseline}\label{lingm}
\end{table}

\begin{table}[t!]
\centering
\begin{tabular}{|l|c|c|c|}
\hline \bf Category & \bf Baseline & \bf SMPI & \bf IMPI \\
\hline
Conditional & \textbf{76.9} & 69.2 & \textbf{76.9} \\
Word overlap & 81.1 & \textbf{89.2} & 86.5 \\
Negation & \textbf{76.0} & 73.1 & 75.0 \\
Antonym & 80.0 & 80.0 & \textbf{85.0} \\
Long Sentence & 73.4 & 69.7 & \textbf{78.0} \\
Tense Difference & \textbf{77.8} & \textbf{77.8} & 72.2 \\
Active/Passive & \textbf{90.0} & \textbf{90.0} & \textbf{90.0} \\
Paraphrase & 86.5 & \textbf{89.2} & \textbf{89.2} \\
Quantity/Time & 64.1 & \textbf{66.7} & 64.1 \\
Coreference & \textbf{89.7} & 82.8 & 82.8 \\
Quantifier & \textbf{77.9} & 72.1 & 76.4 \\
Modal & \textbf{81.7} & 75.4 & 77.0 \\
Belief & \textbf{87.9} & 84.5 & 86.5 \\
\hline
Overall & 77.8 & 78.6 & \textbf{78.9} \\
\hline
\end{tabular}
\caption{Linguistic Error Analysis on MultiNLI development mismatched sets. Baseline is the cross alignment baseline}\label{lingmm}
\end{table}

\begin{figure*}[t]
    \centering
    \includegraphics[width=16cm]{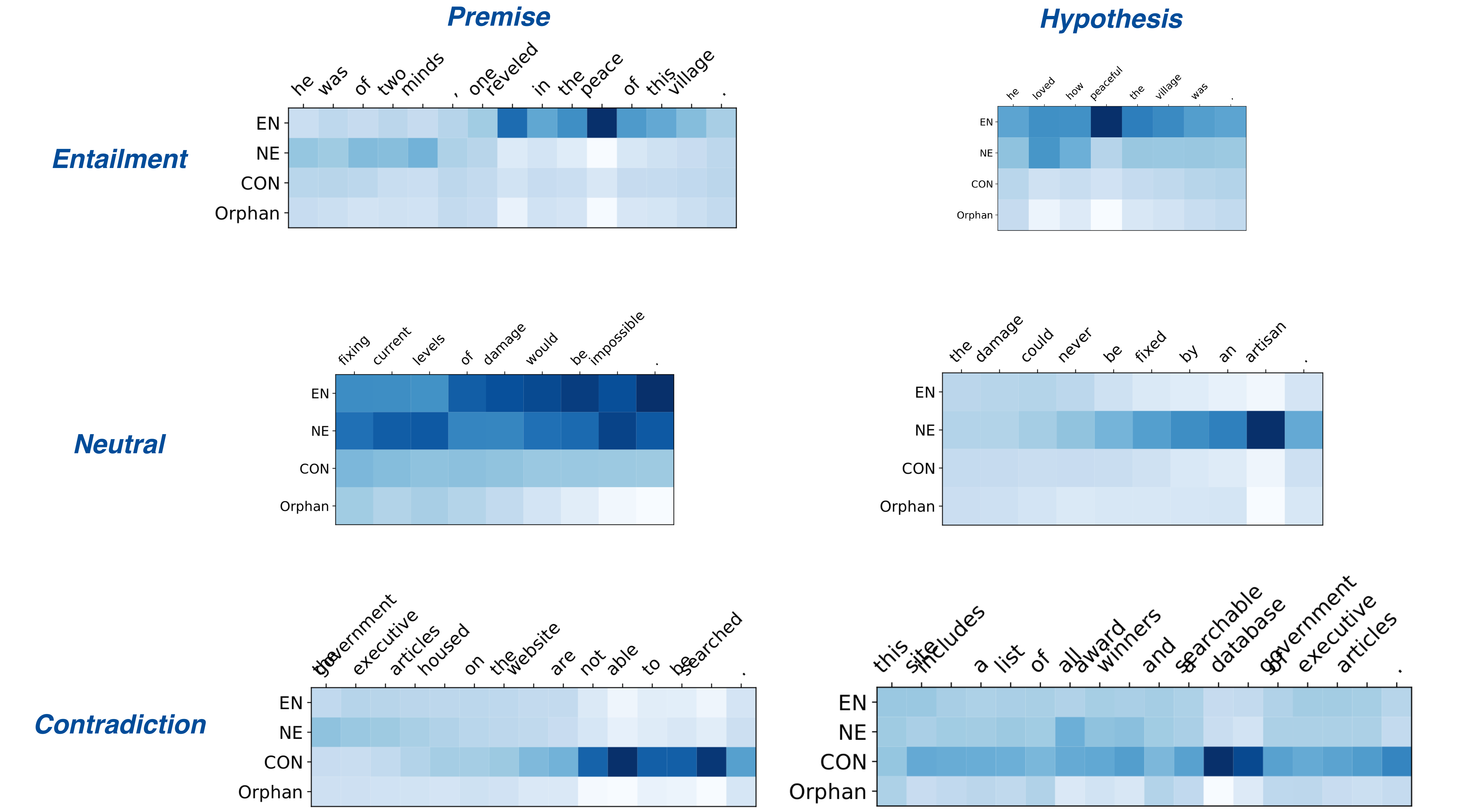}
    \caption{Routing weights $c_{il}$ visulization of SMPI with three samples in MultiNLI development dataset. Darker color correspond to a higher routing weights.}
    \label{visual}
\end{figure*}

\begin{table*}[t]
\centering
\footnotesize
\begin{tabular}{|c|c|c|c|c|c|c|}
\hline \bf Relation & \bf [0,0.2) & \bf [0.2,0.4) & \bf [0.4,0.6) & \bf [0.6,0.8) & \bf [0.8,1.0) & \bf 1.0 \\
\hline
\textsc{En} & 458/317 & 1331/1221 & 986/1123 & 513/599 & 147/161 & 44/42 \\
\textsc{Ne} & 982/849 & 1254/1264 & 495/578 & 253/290 & 68/101 & 71/47 \\
\textsc{Con} & 845/684 & 1368/1459 & 651/725 & 284/289 & 54/75 & 11/8 \\
\hline
Total & 2285/1850 & 3953/3944 & 2132/2426 & 1050/1178 & 269/337 & 126/97 \\
\hline
\end{tabular}
\caption{Statistics of MultiNLI development matched/mismatched sets divided by the word overlap rates into six groups: [0,0.2), [0.2,0.4), [0.4,0.6), [0.6,0.8), [0.8,1.0) and 1.0.}\label{woamount_app}
\end{table*}

\section{Visualization on Routing Assignment Weights}

Here we will show more visualization on routing assignment weights $c_{il}$ in Figure \ref{visual}.

\section{Overall Analysis on Word Overlap Rate}

\begin{figure*}[t] 
  \centering
  \subfigure[Total.] {
    \label{fig:m1}
    \includegraphics[width=5cm]{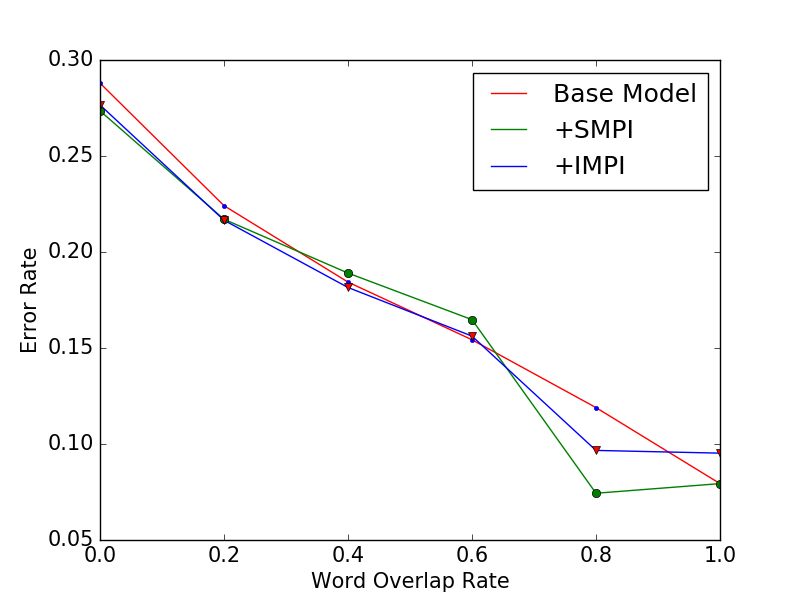}
  }
  \hspace{0.01\linewidth}
  \subfigure[Entailment.] {
    \label{fig:m2}
    \includegraphics[width=5cm]{en_wor_m.png}
  }
  \quad
  \subfigure[Neutral.] {
    \label{fig:m3}
    \includegraphics[width=5cm]{ne_wor_m.png}
  }
  \subfigure[Contradiction.] {
    \label{fig:m4}
    \includegraphics[width=5cm]{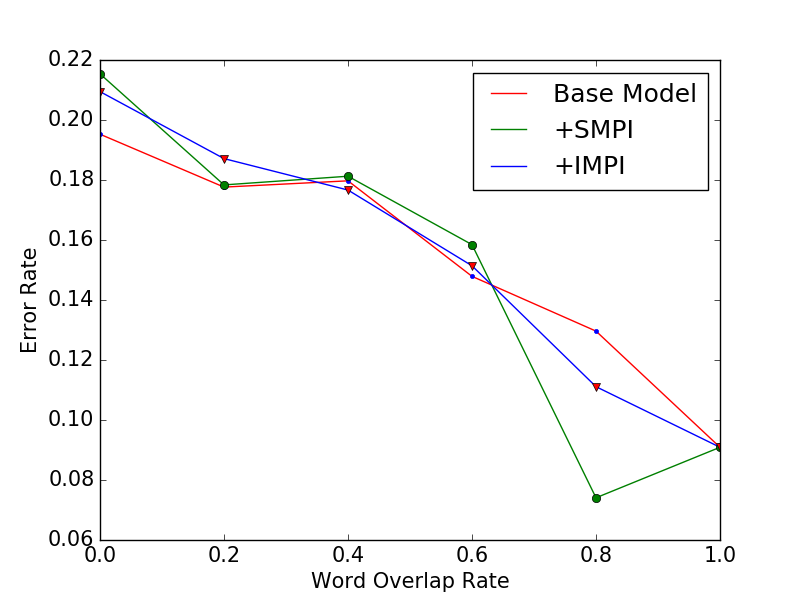}
  }
  \caption{Error rate of total and three inference classes (entailment, neutral and contradiction) by word overlap rate in MultiNLI development matched set. Base Model is the cross alignment baseline. Our MPIs relieve the error rate in the intervals with low word overlap rate of entailment and neutral significantly.} 
  \label{fig:m}
\end{figure*}

\begin{figure*}[t] 
  \centering
  \subfigure[Total.] {
    \label{fig:hyper1}
    \includegraphics[width=5cm]{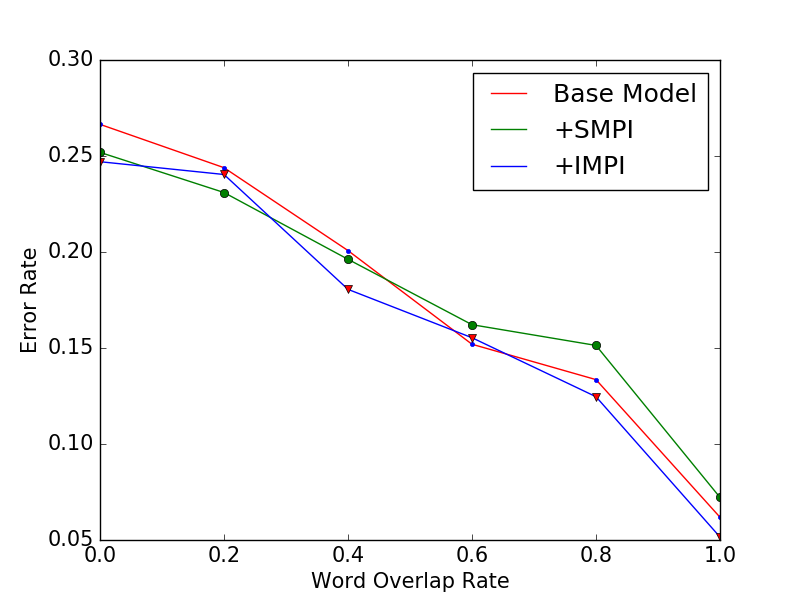}
  }
  \hspace{0.01\linewidth}
  \subfigure[Entailment.] {
    \label{fig:hyper2_app}
    \includegraphics[width=5cm]{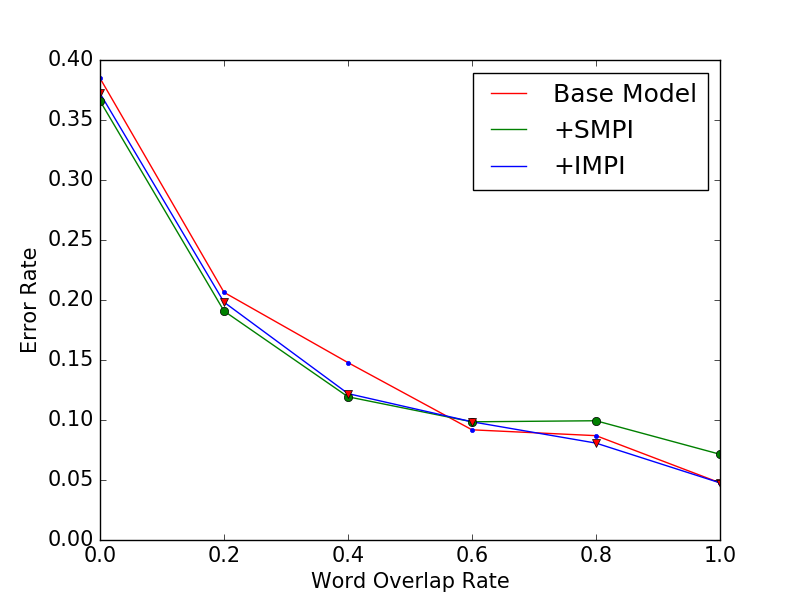}
  }
  \quad
  \subfigure[Neutral.] {
    \label{fig:hyper3_app}
    \includegraphics[width=5cm]{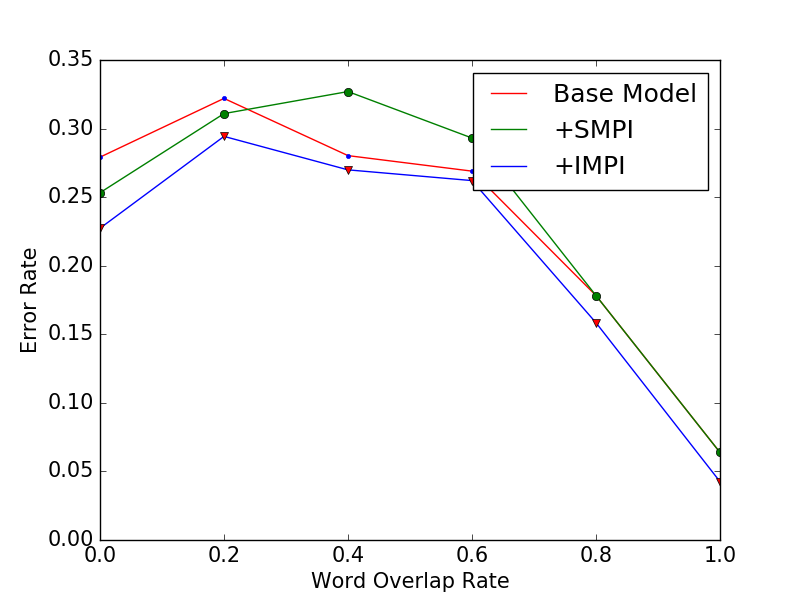}
  }
  \subfigure[Contradiction.] {
    \label{fig:hyper4}
    \includegraphics[width=5cm]{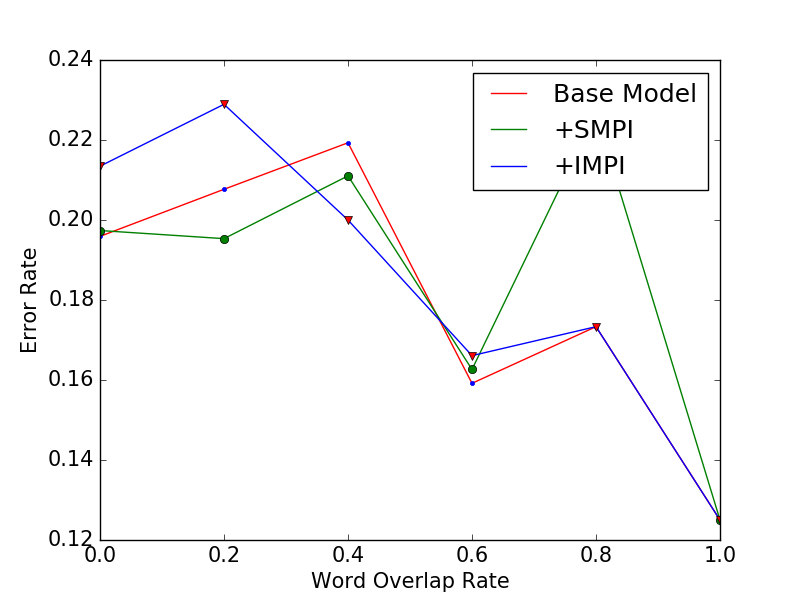}
  }
  \caption{Error rate of total and three inference classes (entailment, neutral and contradiction) by word overlap rate in MultiNLI development mismatched set. Base Model is the cross alignment baseline. Our MPIs relieve the error rate in the intervals with low word overlap rate of entailment and neutral significantly.}
  \label{fig:hyper_app}
\end{figure*}

Table \ref{woamount_app} shows the statistic about the MultiNLI development matched/mismatched sets divided into six groups by word overlap rates: [0,0.2), [0.2,0.4), [0.4,0.6), [0.6,0.8), [0.8,1.0) and 1.0. Figure \ref{fig:m} and \ref{fig:hyper_app} show the corresponding error rates of our base model and two MPIs. Obviously, our MPIs relieve the phenomenon: the small amount aligned parts can cause the misjudgment between entailment relations and neutral relations.
As the Figure \ref{fig:m1} and Figure \ref{fig:hyper1} show, the total error rate is much higher when the aligned parts in the small amount. This phenomenon is more serious in entailment class and neutral class---40\% and 30\%.\footnote{the error rate of low word overlap rate on contradiction is about 20\% which is close to the overall
performance. The judgment of
contradiction does not only need aligned parts (to find the same object like entailment) also need
unaligned parts to find the conflict (somewhat like neutral), word overlap rate is hard to quantify it.} We hope that NLI models should not only focus on capturing the aligned parts, also take advantage of the unaligned parts. The described phenomenon in NLI will be relieve in this way and NLI models can achieve better performance.

\end{document}